\documentclass[10pt,twocolumn,letterpaper]{article}
\usepackage{cvpr}
\usepackage{times}
\usepackage{epsfig}
\usepackage{graphicx}
\usepackage{amsmath}
\usepackage{amssymb}
\usepackage{subcaption}
\usepackage{mathrsfs}
\usepackage{array}
\usepackage{bbm}
\usepackage{authblk}

\usepackage[pagebackref=true,breaklinks=true,letterpaper=true,colorlinks,bookmarks=false]{hyperref}

\cvprfinalcopy % *** Uncomment this line for the final submission

\def\cvprPaperID{765} % *** Enter the CVPR Paper ID here

\ifcvprfinal\fi
\begin{document}

%%%%%%%%% TITLE
\title{Multiple Anchor Learning for Visual Object Detection}

\author[1]{Wei Ke\thanks{indicates equal contributions }}
\author[2]{Tianliang Zhang\footnote[1]}
\author[1]{Zeyi Huang}
\author[2]{Qixaing Ye\thanks{indicates corresponding author}}
\author[3]{Jianzhuang Liu}
\author[1]{Dong Huang}

\affil[2]{Carnegie Mellon University \\
Pittsburgh, US
}
\affil[2]{University of Chinese Academy of Sciences\\
Beijing, China
}
\affil[3]{Shenzhen Institutes of Advanced Technology\\
Shenzhen, China
\authorcr\small \{weik, zeyih\}@andrew.cmu.edu, zhangtianliang17@mails.ucas.ac.cn, jz.liu@siat.ac.cn, donghuang@cmu.edu}

\maketitle
%\thispagestyle{empty}

%-----------------------------------------------------------------------------------------
\begin{abstract}

Classification and localization are two pillars of visual object detectors. However, in CNN-based detectors, these two modules are usually optimized under a fixed set of candidate (or anchor) bounding boxes. This configuration significantly limits the possibility to jointly optimize classification and localization. In this paper, we propose a Multiple Instance Learning (MIL) approach that selects anchors and jointly optimizes the two modules of a CNN-based object detector. Our approach, referred to as Multiple Anchor Learning (MAL), constructs anchor bags and selects the most representative anchors from each bag. Such an iterative selection process is potentially NP-hard to optimize. To address this issue, we solve MAL by repetitively depressing the confidence of selected anchors by perturbing their corresponding features. In an adversarial selection-depression manner, MAL not only pursues optimal solutions but also fully leverages multiple anchors/features to learn a detection model. Experiments show that MAL improves the baseline RetinaNet with significant margins on the commonly used MS-COCO object detection benchmark and achieves new state-of-the-art detection performance compared with recent methods. 
% \footnote{The code is available at \href{https://github.com/CVPR765}{\color{magenta}github.com/CVPR765.}}
\end{abstract}

%-----------------------------------------------------------------------------------------
\section{Introduction}

\begin{figure}
  \centering
  \begin{subfigure}[b]{0.514\linewidth}
      \includegraphics[width=\linewidth]{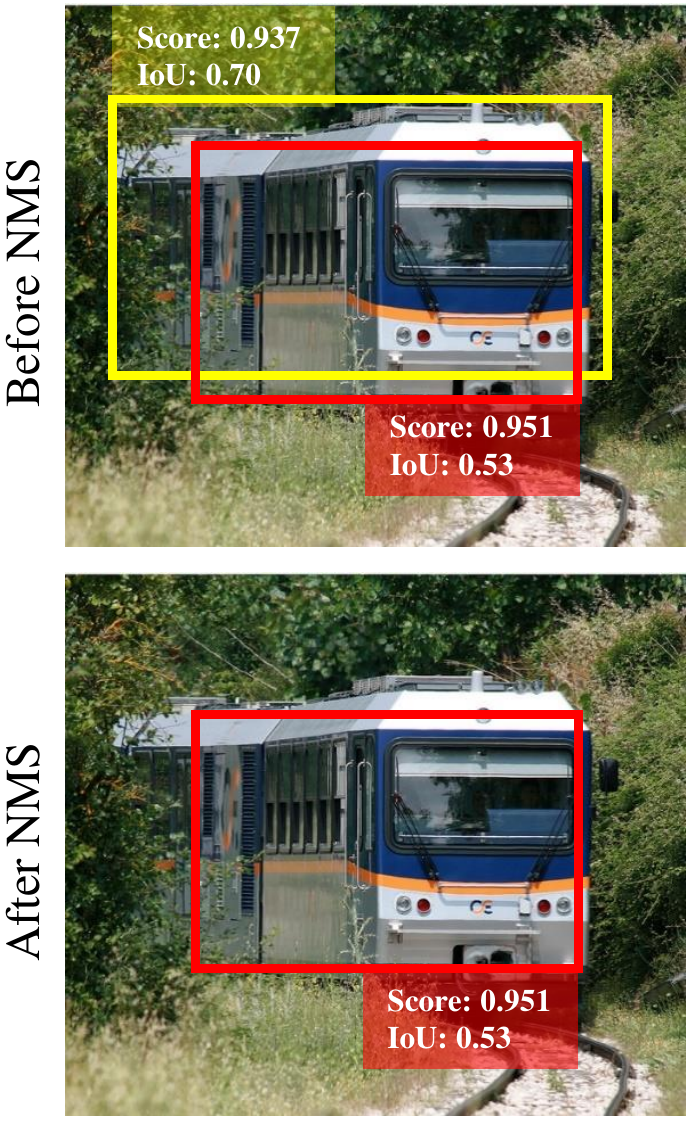}
      \caption{Baseline}
      \label{fig:fig1a}
  \end{subfigure}
  \begin{subfigure}[b]{0.47\linewidth}
      \includegraphics[width=\linewidth]{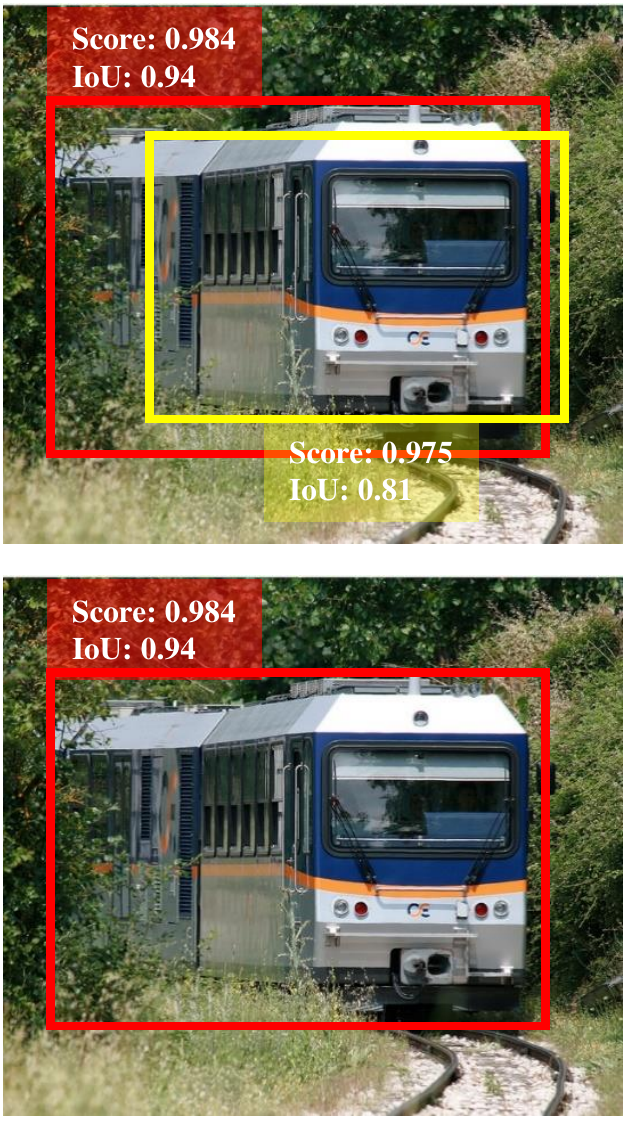}
      \caption{MAL}
      \label{fig:fig1b}
  \end{subfigure}
    \caption{Detection outputs of the baseline detector (RetinaNet) and the Multiple Anchor Learning (MAL), before and after NMS. The baseline detector may produce bounding boxes with high localization IoU with a low classification score (the yellow bbox), or low localization IoU with a high classification score (the red bbox), which lead to sub-optimal results after NMS. MAL produces bounding boxes with high co-occurrence of top  classification  and  localization, leading to better detection results after NMS. }
  \label{fig:moti}
\end{figure}

Convolutional Neural Network (CNN) based object detectors have achieved unprecedented advances in the past few years ~\cite{RCNN14,FastRCNN15,FasterRCNN15,YOLO16,SSD16,FPN17,FocalLoss17}. In both recent two-stage and single-stage object detectors, the bounding box classification and localization modules are highly integrated: they are conducted on the shared local features, and are optimized over the sum of the loss functions.  

To provide rich candidates of shared local features, a prevalent approach is the introduce hand-crafted dense anchors~\cite{FocalLoss17} on the convolutional feature maps. These anchors create a uniform distribution of bounding box scales and aspect ratios, enabling objects with various scales and aspect ratios to be equally represented in training a detector. 

However, optimization under a fixed set of hand-crafted anchors significantly limits the possibility to jointly optimize classification and localization. During training, detectors leverage spatial alignment, $i.e.$, Intersection over Unit (IoU) between objects and anchors, as the sole criterion to assign anchors. Each assigned anchor independently supervises network learning for classification and localization. Without direct interactions of the two optimizations, the detections of accurate localization may have lower classification confidence, and be suppressed by the following Non-Maximum Suppression (NMS) procedure (see the baseline example in Fig.~\ref{fig:moti}).

Recent remedy for the problem includes IoU-Net~\cite{IoU-Net18} and FreeAnchor~\cite{zhang2019freeanchor}. However, it remains using independent classification and localization confidence during the training procedure. FreeAnchor selects anchors according to a joint probability over classification and localization. Nevertheless, the matching procedure based on maximum likelihood estimation (MLE) is not optimal considering the non-convexity of the problem.

\begin{figure*}[t]
\begin{center}
   \includegraphics[width=\linewidth]{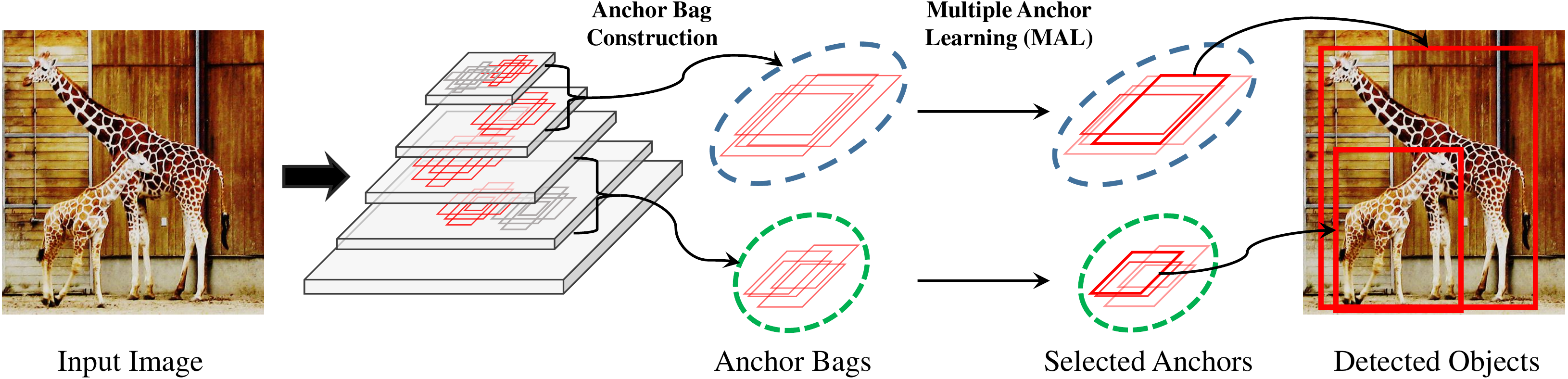}
\end{center}
\caption{The main idea of MAL. In the feature pyramid network, an anchor bag $A_i$ is constructed for each object $b_i$. Together with the network parameter learning, $i.e.,$ back-propagation, MAL evaluates the joint classification and localization confidence of each anchor in $A_i$. Such confidence is used for anchor selection and indicates the importance of anchors during network parameter evolution.}
\label{fig:MAL}
\end{figure*}

In this paper, we present Multiple Anchor Learning (MAL), an automatic anchor learning approach that jointly optimizes object classification and localization from the perspective of anchor-object matching. In training phase of MAL, an anchor bag for each object is constructed by choosing the top ranked anchor with IoUs between anchors and the object bounding box.
MAL evaluates positive anchors in each bag by combining their classification and localization scores. In each training iteration, MAL uses all positive anchors to optimize the training loss but selects the high/top-scored anchors as the solutions. This leads to high co-occurrence of top classification and localization (see the MAL example in Fig.~\ref{fig:MAL}). 

MAL is optimized over an anchor selection loss based on Multiple Instance Learning (MIL)~\cite{MIL97}. However, the iterative selection process under conventional MIL is potentially NP-hard to optimize. Selecting the top-scored instance (anchor) in each learning iteration could produce sub-optimal solutions, $e.g.,$ falsely localized object parts. To address this issue, we solve MAL by repetitively depressing the confidence of top-scored anchors by perturbing their features, which guarantees that potential optimal solutions, $i.e.$, positive anchors of lower confidence, have an opportunity to participate in learning. By upgrading the supervision from independent anchors to multiple anchors, MAL fully leverages multiple anchors/features to learn a better detector. The contributions of this work include:

\begin{itemize}
    \vspace{-0.8em}
    \item  We propose an Multiple Anchor Learning (MAL) approach, jointly optimizing classification and localization modules for object detection by evaluating and selecting anchors. 
    \vspace{-0.8em}
    \item  We propose a selection-depression optimization strategy, providing an elegant-yet-effective way to prevent MAL from getting stuck into sub-optimal solutions during detector training.
    \vspace{-0.8em}
    \item We improve state-of-the-arts with significant margins on the commonly used MS COCO dataset.  

\end{itemize}

%-----------------------------------------------------------------------------------------
\section{Related Work}

Various taxonomies~\cite{Survey2019} have been used to categorize the large amount of CNN-based object detection methods, $e.g.,$ one-stage~\cite{FasterRCNN15} vs. two-stage~\cite{FocalLoss17}, single-scale features ~\cite{FasterRCNN15} vs. multi-scale representation~\cite{FPN17,Triplet2019,ScaleSensitive2019}, and handcrafted architectures~\cite{SSD16} vs. Network Architecture Search (NAS)~\cite{NAS-FPN2019}. In this paper, we review the related works from the perspective of object localization.

\subsection{Anchor-Based Method} 
Training a detector requires to generate a set of bounding boxes along with their classification labels associated with the objects in an image. 
However, it is not trivial for CNNs to directly predict an order-less set of arbitrary cardinals~\cite{MetaAnchor2018}. One commonly used strategy is to introduce anchors, which employs a divide-and-conquer strategy to match objects with convolutional features, spatially.

\textbf{Anchor Assignment.}
Anchor-based detection methods include the well-known Faster R-CNN~\cite{FasterRCNN15}, FPN~\cite{FPN17}, RetinaNet~\cite{FocalLoss17}, SSD~\cite{SSD16}, DSSD~\cite{Fu2016dssd}, and YOLO~\cite{YOLO9000}. 
In these detectors, a large amount of anchors are scattered over convolutional feature maps so that they can match objects of various aspect ratios and scales. 
During training, the anchors are assigned to objects (positive anchors) or backgrounds (negative anchors)  by threshold their IoUs with the ground-truth bounding boxes~\cite{FasterRCNN15}. During inference, anchors independently predict object bounding boxes, where the box with the highest classification score is retained after the NMS procedure.

Despite of the simplicity, these approaches rely on the assumption that anchors are optimal for both object classification and localization. For objects of partially occlusion and irregular shapes, however, such heuristics are implausible and they could miss the best anchors/features~\cite{zhang2019freeanchor}.  

\textbf{Anchor Optimization.} To pursue optimal feature-object matching, MetaAnchor~\cite{MetaAnchor2018} learns to predict anchors from the arbitrary customized prior boxes with a subnet. GuidedAnchoring~\cite{GuidedAnchoring} leverages semantic features to guide the prediction of anchors while replacing dense anchors with predicted anchors.
FreeAnchor~\cite{zhang2019freeanchor} upgrades handcrafted anchor assignment to ``free" anchor matching. This approach formulates detector training as a maximum likelihood estimation (MLE) procedure. Its goal is to learn features that best explain a class of objects in terms of both classification and localization. 
IoU-Net~\cite{IoU-Net18} selects anchors while predicting the IoU between a detected bounding box and a ground-truth box.  
Combined with an IoU-guided NMS, IoU-Net reduces the suppression failure caused by the misleading classification confidences.
Gaussian YOLO~\cite{GaussianYOLO2019} introduces localization uncertainty that indicates the reliability of anchors/bounding boxes. By using the estimated localization uncertainty during inference, this approach improves classification and localization accuracy. 

All above approaches have taken some steps towards anchor learning. Nevertheless, how to efficiently select optimal anchors remains to be further elaborated. Considering a non-convex objective function which could cause sub-optimal solutions, we propose an adversarial selection-depression strategy to alleviate this issue.

%to the best of our knowledge, there still lacks an approach for joint and consistent anchor evaluation in both training and inference procedures, which inhibits the optimization of feature selection and feature learning. 

\subsection{Anchor-Free Method}
Instead of using anchors as bases to conduct detection, researchers have recently explored anchor-free approaches, which operates on individual cells of the convolutional feature maps.
%
% EAST detector~\cite{East2017} \textcolor{red}{uses all feature cells with the object scores to learn detection models}, while selecting a single box of the highest classification score for object localization. % EAST is for text detection
%
FCOS leverages cell-level supervision and center-ness bounding-box regression~\cite{tian2019fcos} for object detection. CornerNet~\cite{CornerNet2018} and CenterNet~\cite{CenterNet2019} replace bounding box supervision with key-point supervision. Extreme point~\cite{ExremePoint2019} and RepPoint~\cite{RepPoint2019} use point sets to predict object bounding boxes.   

As a new direction for object detection, anchor-free methods show great potential for extreme object scales and aspect ratios, without constraints set by hand-craft anchors. However, without the anchor box as the reference point, direct regression of bounding boxes from convoltuional features remains a very challenging problem. As an anchor-based approach, MAL outperforms the current top anchor-free detectors such as CenterNet and CornerNet. 

\section{The Proposed Approach}

MAL is implemented based on RetinaNet ~\cite{FocalLoss17} network architecture. MAL upgrades RetinaNet by finding optimal selection of anchors/features for both classification and localization. In what follows, we briefly revisit RetinaNet on its original mechanism in object classification and localization. We then elaborate how MAL improve classification and localization by evaluating anchors. We finally propose an anchor selection-depression strategy to pursue optimal solutions of MAL. 
%Our description is about one feature layer. For multi-layer features in RetinaNet, MAL is implemented identically on each layer.
%not only explcicitly pursue the optimal selection of anchors but also explicitly implements feature ensemble during the detector training procedure. 
%alleviate the sample unbalance issue {\color{red}persists} in detector training in single-stage object detection framework. 

\begin{figure*}[t]
\begin{center}
   \includegraphics[width=1\linewidth]{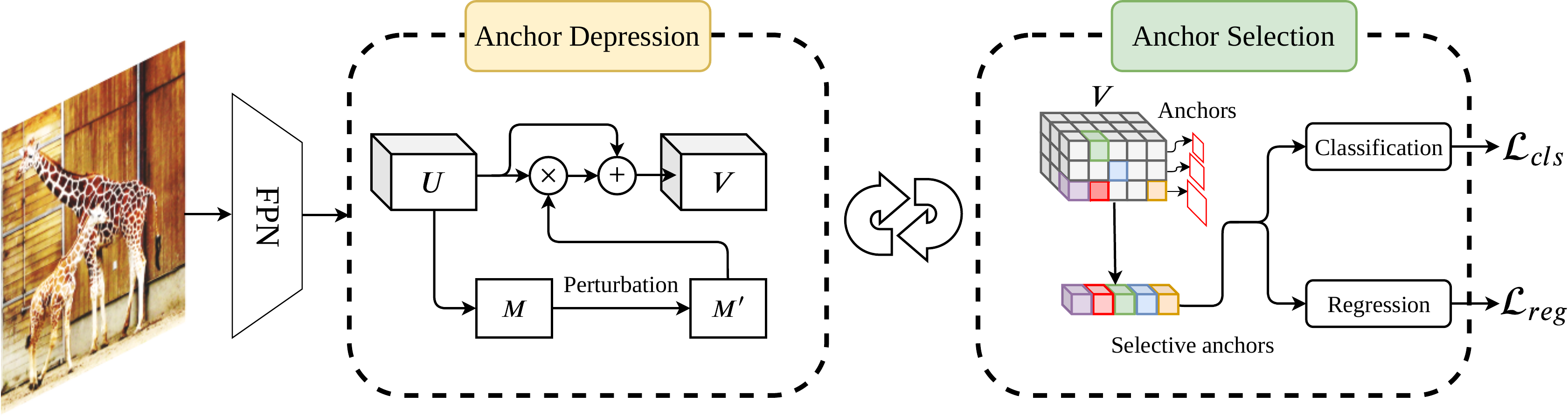}
\end{center}
\caption{MAL implementation. During training, it includes the additional anchor selection and anchor depression modules added to RetinaNet. During test, it uses exactly the same architecture as RetinaNet. ``$U$" and ``$V$" respectively denote convolutional feature maps before and after depression. ``$M$" and ``$M'$" respectively denote an activation map before and after depression.}
\label{fig:impl}
\end{figure*}

%-----------------------------------------------------------------------------------------
\subsection{RetinaNet Revisit}
RetinaNet is a representative architecture of single-stage detectors with state-of-the-art performance. A RetinaNet detector is made up of a backbone network and two subnets, one for object classification and other for object localization. Feature Pyramid Network (FPN) is used at the end of RetinaNet backbone network. From each feature map in the feature pyramid, a classification subnet predicts category probabilities while a box regression subnet predicts object locations using anchor boxes as the reference locations. The input features of the two subnets are shared across the feature pyramid levels for efficiency. Considering the extreme imbalance of foreground-background classes, presented as positive-negative anchors after anchor-object matching, Focal Loss is adopted to prevent the vast number of easy negatives from overwhelming the detector during training.

Let $x \in \mathcal{X}$ be an input image with label $y \in \mathcal{Y}$, where $\mathcal{X}$ is the training image set and $\mathcal{Y}$ is the label set of the categories.
% {\color{red} $\mathcal{Y}=\{1, 0\}$ indicates whether $x$ contains positive samples or not.??} 
%
Without loss of generality, denote $B$ as the ground-truth bounding boxes of the objects in a positive image. $b_i\in B$ consists of the class label $b_i^{cls}$ and the spatial position $b_i^{loc}$.
%while $A$ denote the anchor set for an image.
% $a_j$ denotes an anchor, for which the classification confidence $a_j^{cls}$ and bounding-box output $a_j^{loc}$ of $a_j$ are predicted by the classification subnet and box regression subnet, respectively.
The classification confidence $a_j^{cls}$ and bounding-box output $a_j^{loc}$ of the anchor $a_j$ are predicted by the classification and the box regression subnets, respectively.
The anchors in an image are divided into positive ones $a_{j+}$ if their IoUs with the ground-truth boxes are larger than a threshold, and negative ones $a_{j-}$ otherwise. Anchors are used to supervise network learning, as 
\begin{equation}
    \theta^*= \arg\max_{\theta} \big(f_{\theta}(a_{j+}, b_i^{cls})-\gamma f_{\theta}(a_{j-}, b_i^{cls})\big),
    \label{eq.cls}
\end{equation}
where $f_{\theta}(\cdot)$ denotes the classification procedure, and $\gamma$ is a factor to balance the importance of negative/positive anchors. Simultaneously, positive anchors are used to optimize the object localization, as
\begin{equation}
    \theta^*= \arg\max_{\theta} g_{\theta}(a_{j+}, b_i^{loc}),
    \label{eq.loc}
\end{equation}
where $\theta$ denotes the network parameters, and $g_{\theta}(\cdot)$ denotes the bounding-box regression procedure. Eq.\ \ref{eq.cls} and  Eq.\ \ref{eq.loc} are actually implemented by minimizing the Focal Loss, $\mathcal{L}_{cls}(a_j, b_i^{cls})$, and the Smooth-L1 loss, $\mathcal{L}_{loc}(a_j, b_i^{loc})$, 

During network learning, each assigned anchor independently supervises the learning for object classification and object localization, without considering whether the detection and localization are compatible on assigned anchors. This could cause the anchors of accurate localization but with lower classification confidence to be suppressed by the following Non-Maximum Suppression (NMS) procedure. 

%-----------------------------------------------------------------------------------------
\subsection{Multiple Anchor Learning}

To alleviate the drawbacks of independent anchor optimization, we propose the Multiple Anchor Learning (MAL) approach, Fig.\ \ref{fig:MAL}. % inspired by MIL optimization procedure. 
In each learning iteration, MAL selects high-scored instances in an anchor bag to update the model. 
% The updated model is in turn to evaluate each instance with new confidence. 
After updating, the model evaluates each instance with new confidence. 
Model learning and anchor selection iteratively perform towards final optimization.

% To fulfill this purpose, we first construct an anchor bag $A_i$ for the $i^{th}$ object which are the top-$k$ ones considering the IoU between them and ground truth.
To fulfill this purpose, we first construct an anchor bag $A_i$ for the $i^{th}$ object. The anchor bag includes the top-$k$ anchors according to the IoUs between the anchors and the ground truth.
Together with network parameter learning, $i.e.,$ back-propagation, MAL evaluates the joint classification and localization confidence of each anchor in $A_i$. Such confidence is used for anchor selection and indicates the importance of anchors during network parameter evolution. For simplicity, consider solely the learning upon positive anchors, while that for negative anchors follows Eq.\ \ref{eq.cls}. MAL has the following objective function:
\begin{equation}
\begin{split}
    \{\theta^*,a_i^*\} &= {\arg\max}_{\theta,a_j \in A_i}  F_{\theta}(a_j, b_i)\\
    &={\arg\max}_{\theta, a_j \in A_i} f_{\theta}(a_j, b_i^{cls})+\beta g_{\theta}(a_j, b_i^{loc}),\\
    %&-\sum_{\theta, a_j \in A_-} f_{\theta}(a_j, b_i^{cls})
\end{split}
    \label{eq.loc-reg}
\end{equation}
where $f_{\theta}(.)$ and $g_{\theta}(.)$ give the classification and localization scores, respectively, and $\beta$ is a regularization factor. It is towards selecting a best positive anchor $a_i^*$ for the $i^{th}$ object, as well as learning the network parameters $\theta^*$.  

The objective function defined in Eq.\ \ref{eq.loc-reg} is converted to a loss function as:
\begin{equation}
\begin{split}
    \{\theta^*,a_i^*\} &= {\arg\min}_{\theta,a_j \in A_i}{\mathcal{L}_{det}(a_j, b_i)} \\
    &= {\arg\min}_{\theta,a_j \in A_i} {\mathcal{L}_{cls}(a_j, b_i^{cls})} + \beta {\mathcal{L}_{reg}(a_j, b_i^{loc})},
 \end{split}
 \label{eq.loc-reg-loss}
\end{equation}
where $\mathcal{L}_{cls}$ and $\mathcal{L}_{reg}$ are the classification and detection losses, respectively, as defined in Section 3.1. The loss for negative anchors follows the Focal Loss defined in~\cite{FocalLoss17}. 

%where $a_j^{cls*}$ and $a_j^{loc*}$ is the label and position of the corresponding ground-truth. The $\mathcal{L}_{cls}$ is cross entropy loss and $\mathcal{L}_{reg}$ is $smoothL1$ loss, where $\mathcal{L}_{reg} = 0$ when $a_j$ is background.

%\begin{equation}
%    \mathcal{L}= \sum_{j} {\mathcal{L}_{cls}(a_j^{cls}, a_j^{cls*})} + \beta {\mathcal{L}_{reg}(a_j^{loc}, a_j^{loc*})}
%\end{equation}
%where $a_j^{cls*}$ and $a_j^{loc*}$ is the label and position of the corresponding ground-truth. The $\mathcal{L}_{cls}$ is cross entropy loss and $\mathcal{L}_{reg}$ is $smoothL1$ loss, where $\mathcal{L}_{reg} = 0$ when $a_j$ is background.
%It's rewritten as:
%\begin{equation}
%    w^*= arg \ max \sum_{a_j \in A}f(a_j, w).
%\end{equation}
%where $w$ is the network parameters and $f(a_j, w) = - CE(a_j^{cls}, a_j^{cls*}) * \beta SmoothL1(a_j^{loc}, a_j^{loc*})$ is a confidence function for an anchor with the detector, which minimizes the loss in Eq. (1). The network parameters are affected by all matched anchors equally during early training epoch while by the highest scored anchor during the later epochs.

%-----------------------------------------------------------------------------------------
\subsection{Selection-Depression Optimization}
\label{sect:seS-D}

%Considering MIL's non-convexity, 
%
% To optimize Eq.\ \ref{eq.loc-reg} and Eq.\ \ref{eq.loc-reg-loss}, it requires to learn, at the same time, anchors and network parameters. 
% This is a non-convex problem and MAL could cause sub-optimal anchor selection. 
Optimizing Eq.\ \ref{eq.loc-reg} or Eq.\ \ref{eq.loc-reg-loss} with Stochastic Gradient Descent (SGD) is a non-convex problem, which could cause a sub-optimal anchor selection. To alleviate the problem and select optimal anchors, we propose repetitively depressing the confidence of selected anchors by perturbing their corresponding features. Such a learning strategy, referred to as selection-depression optimization, solves the MAL problem in an adversarial manner. 
%not only pursues optimal solutions but also fully leverages multiple anchors/features to learn a detection model.
%The selection is solve the shortage to RetinaNet discussed above, while depression perturbs the selected anchors who gets the highest confidence score. By this adversarial manner, it alleviate the sub-optimal anchor selection. MAL not only guarantees the optimal selection of anchors but also implements implicit feature ensemble by assigning anchors/feature different  confidence.
% Eq.\ \ref{eq.loc-reg-loss} defines a loss function to update the network parameters. 
% In each learning iteration, using the updated network parameters, we calculate the confidence of anchor $a_j$ corresponding to object $b_i$ using $F_{\theta}(a_j, b_i)$ defined in Eq.\ \ref{eq.loc-reg}. According to $F_{\theta}(a_j, b_i)$, the conventional MIL algorithm tends to select the top-scored anchor. Nevertheless, in the context of object detection, selecting a top-scored anchor from each bag is implausible, as validated by the continuation MIL method~\cite{cvpr_WanLKJJY19}. The reason lies in the non-convex objective function of MIL, which could introduce sub-optimal solutions considering that the discriminative capability of network is very limited during early iterations. To alleviate this issue, we proposed the following ``all-to-top-1" anchor selection strategy.

%From Eq. (1), RetinaNet is not optimized effectively when the highest scored anchor is not the best matched one. It motivated us to use MIL to select anchor and learn network parameters.

\textbf{Anchor Selection.} According to $F_{\theta}(a_j, b_i)$, the conventional MIL algorithm tends to select the top-scored anchor. Nevertheless, in the context of object detection, selecting a top-scored anchor from each bag is difficult, as validated by the continuation MIL method~\cite{cvpr_WanLKJJY19}. Instead of selecting the highest-scored anchor in Eq. \ref{eq.loc-reg} in the training phase, we propose an ``All-to-Top-1" anchor selection strategy from each anchor bag for back-propagation. When learning proceeds, we linearly decrease the number from $|A_i|$ (number of anchors in a bag) to 1. 
% Supposing $\lambda = t/T$ where $t$ and $T$ are current and total numbers of iteration for training, and $\phi(\lambda)$ which indicates the indices of top-ranked anchors. $|\phi(\lambda)| = |A_i|*(1-\lambda)+1$, Accordingly, Eq. \ref{eq.loc-reg} is re-written as:
Formally, let $\lambda = t/T$, where $t$ and $T$ are the current and total numbers of iterations for training. Then let $\phi(\lambda)$ indicate the indices of high-ranked anchors and $|\phi(\lambda)| = |A_i|*(1-\lambda)+1$. Finally, Eq.~\ref{eq.loc-reg} is re-written as:
\begin{equation}
    \{\theta^*,a_i^*\} = {\arg\max}_{\theta,a_j \in A_i} \sum_{j \in \phi(\lambda)} F_{\theta}(a_j, b_i),
    \label{eq.loc-reg-re}
\end{equation}
%It means that all positive anchors in a bag participate network learning at the first learning epoch while only the top-1 anchor is selected in the last epoch. 
Along this pipeline, MAL leverages multiple anchors/features within the object region to learn a detection model in early training epochs, and converges to use a single optimal anchor at the last epoch. 
%

%Specifically, we defines a continuous indicator function $J_s(\lambda)$ to c When the training begin, $\lambda = 0.0$ and all the anchors in the bag are selected. While $\lambda = 1.0$ at the end of training and only the top-scored anchor is selected from the bag. 

%The objective function is written as:
%\begin{equation}
%    (w^*, a^*)= \arg\max \sum_{a \in A_i, A_i \subset  A}f(a, w).
%\end{equation}
%With Eq. (3), we jointly optimize classification for detector parameter $w$ and localization %modules for anchor selection in anchor bag.

%As the $a^{cls}$ and $a_{loc}$ share the same feature of the anchor, we define
%\begin{equation}
%    \begin{matrix}
%    a^{cls} = S^{cls}(v) \\
%    a^{loc} = S^{loc}(v)
%    \end{matrix},
%\end{equation}
%where $v$ is the feature vector of the anchor $a$, $S^{cls}(.)$ and $S^{loc}(.)$ are the selection functions to predict the classification score and regress the localization.

\textbf{Anchor Depression.} %The depression component is under the guidance of number of selected anchors $K$, which indicates the depress level of the feature of the anchor. 
Inspired by the inverted attention network \cite{hzeIAN2019}, we developed an anchor depression procedure to perturb the features of selected anchors in order to decrease their confidences (see more Fig.~\ref{fig:impl}). The rational is to endow unselected anchors with extra chances to participate the training. 
% As shown in Fig.\ \ref{fig:impl}, for feature map $U$, the attention map is computed as $M=\sum_{l} w_l * U_l$, where $w$ is the global average pooling of $U$, $l$ is the channel index of $U$. 
 Formally, we denote the feature map and the attention map as $U$ and $M$, where $M$ is computed as $M=\sum_{l} w_l * U_l$, with $w$ being the global average pooling of $U$ and $l$ being the channel index of $U$. We then generate a new depressed attention map $M' = (1-\mathbbm{1}_{P})*M$ by cutting down the high values to zero, where $\mathbbm{1}$ is the 0-1 indicator function. and $P$ is the high-value position. The feature map is perturbed as:
\begin{equation}
    V=(\textbf{1}+M') \circ U_l , 
    \label{eq.frame-dep}
\end{equation}
where $\textbf{1}$ is the identity matrix and $\circ$ denotes the element-wise multiplication. With the continuation strategy, the depression in Eq. \ref{eq.frame-dep} is reformulated as: 
\begin{equation}
    V=(\textbf{1}+(1-\mathbbm{1}_{\psi(\lambda)})*M) \circ U_l , 
    \label{eq.frame-dep-re}
\end{equation}
where $\psi(\lambda)$ indicates how many pixels to be perturbed.

%  {\color{red} We then randomly perturb the feature maps $V$ corresponding to selected anchors on the activation map.} By perturbing the activation map, we generate a new depressed attention map. The feature is updated by element-multiplying with the depressed attention map in a pixel-wise manner. 

\begin{figure}[t]
\begin{center}
   \includegraphics[width=0.9\linewidth]{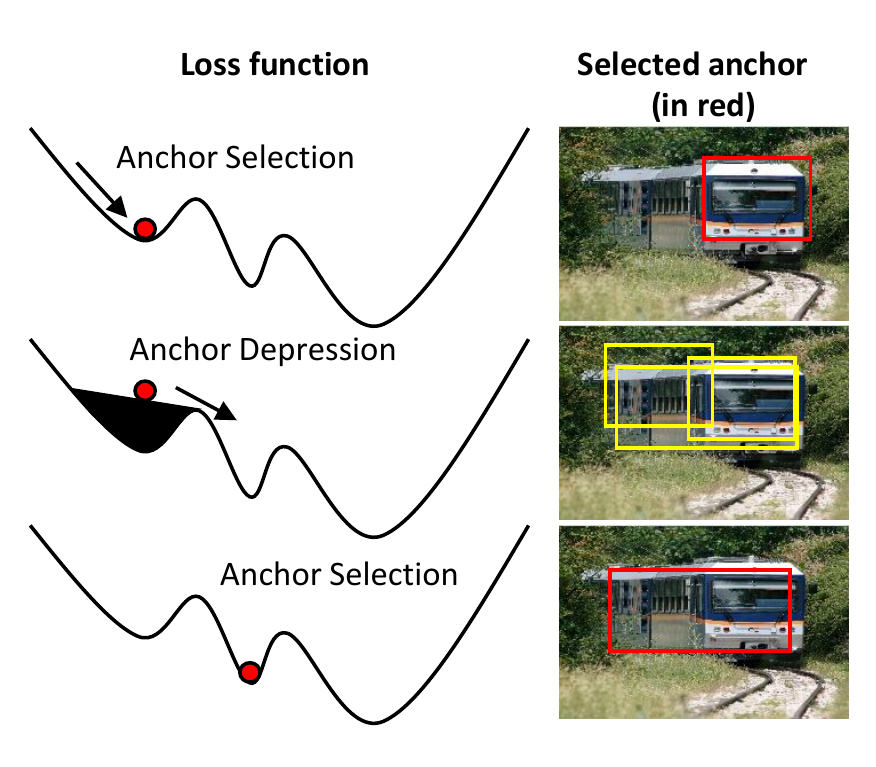}
\end{center}
\caption{Optimization analysis. In the first curve, MAL selects a sub-optimal anchor and gets stuck into a local minimum. In the second curve, anchor depression increases the loss so that MAL continues the optimization. In this way, MAL has a greater chance to find optimal solutions.}
\label{fig:opt}
\end{figure}

%-----------------------------------------------------------------------------------------
\subsection{Implementation}

The implementation of an MAL detector is based on the RetinaNet detector where the features of the input image are extracted by a FPN backbone \cite{FPN17}.
% , detailed in Fig.\ \ref{fig:impl}. 
%
The anchor generation settings are the same as those of RetinaNet, $i.e.$, 9 anchors with three sizes $\{2^0, 2^{1/3}, 2^{2/3}\}$ and three aspect ratios $\{1:2, 1:1, 2:1\}$ for each pixel on the feature maps. Across the levels, the anchors cover the scale range from 32 to 813 pixels with respect to the input image. 

During the feed-forward procedure of the network training, we calculate the detection confidence of each anchor, $F_{\theta}(a_j, b_i)$, to minimize the detection loss defined in Eq.\ \ref{eq.loc-reg-loss}. According to the confidence, top-$k$ anchors are selected. The network parameters are then updated under the supervision of the selected anchors. After anchor selection, anchor depression is carried out, as described in Section\ \ref{sect:seS-D}. In the next iteration, anchor selection is carried out again to select high-scored anchors. 

The inference procedure of our approach is exactly the same as RetinaNet, $i.e.$, we use the learned network parameters to predict classification scores and object bounding boxes, which are fed to a NMS procedure for object detection. As MAL is only applied in the detector training procedure to learn more representative features, our learned detector achieves performance improvement with negligible additional computation cost. 

\subsection{Optimization Analysis}
The anchor selection-depression strategy approximates an adversarial procedure. The selection operation finds top-scored anchors that minimize the detection loss $\mathcal{L}_{det}$. The depression operation perturbs the corresponding features of selected anchors so that their confidence decreases and the detection loss increases again. The selection-depression strategy helps the learner find better solutions for the non-convex objective function of MAL. As illustrated by the first curve of Fig.~\ref{fig:opt}, MAL selects a sub-optimal anchor and gets stuck into a local minimum of the loss function. In the second curve, the anchor depression increases the loss, so that the local minimum is ``filled". Consequently, MAL continues to find the next local minimum. After learning converges, MAL has a better chance to find optimal solutions.

% \begin{table*}[t]
% \centering
% \fontsize{10}{12}\selectfont
% \begin{tabular}{c|c|ccccccc}
% \hline
% Method  & Backbone  & $\mathrm{AP}$   & $\mathrm{AP_{50}}$ & $\mathrm{AP_{75}}$ & $\mathrm{AP_S}$  & $\mathrm{AP_{M}}$  & $\mathrm{AP_{L}}$   \\ \hline \hline
% RetinaNet (baseline)   & ResNet-50       & 35.46 & 51.61 & 	39.37 & 	20.63 & 	39.79 & 	47.19      \\
% MAL+S(all)   & ResNet-50          & 38.14 & 	56.81 & 	40.81 & 	20.12 & 	41.25 & 	52.39   \\
% MAL+S(all-top1)   & ResNet-50     & 38.39 & 	56.81 & 	41.14 & 	20.12 & 	41.97 & 	52.01   \\\hline
% \end{tabular}
% \caption{Comparison of anchor selection functions on COCO minval dataset. ``S" denotes ``Selection".}
% \end{table*}

% \begin{table*}[h]
% \centering
% \fontsize{10}{12}\selectfont
% \begin{tabular}{c|c|ccccccc}
% \hline
% Method       & Backbone      & $\mathrm{AP}$   & $\mathrm{AP_{50}}$ & $\mathrm{AP_{75}}$ & $\mathrm{AP_S}$  & $\mathrm{AP_{M}}$  & $\mathrm{AP_{L}}$   \\ \hline \hline
% RetinaNet (baseline)   & ResNet-50   & 35.46 & 51.61 & 	39.37 & 	20.63 & 	39.79 & 	47.19      \\
% % RetinaNet (baseline) & ResNet-50     & 90k & 35.60 & 	51.89 & 	39.05 & 	19.83 & 	39.94 & 47.42   \\
% MAL+D(constant)  & ResNet-50     & 35.25 & 	51.72 &	38.92 & 19.56 & 	40.32 & 47.99 \\
% MAL+D(step) & ResNet-50     & 35.88 & 	52.34 & 39.63 & 	19.52 & 	40.13 & 	47.61 \\
% MAL+D(symmetric step)  & ResNet-50      &  36.18 & 	52.66 & 	39.88 & 	20.35 & 	40.32 & 	48.67  \\ \hline
% \end{tabular}
% \caption{Comparison of depression strategies on COCO minval dataset. ``D" denotes ``Depression".}
% \end{table*}

\section{Experiments}

In this section, we present experimental results of the proposed Multiple Anchor Learning approach on the bounding-box detection track of the challenging COCO benchmark~\cite{DBLP:conf/eccv/LinMBHPRDZ14}. We follow the common practice and use $\sim$118k images for training, 5k for validation and $\sim$20k for testing without provided annotations (test-dev). 
AP is computed over ten different IoU thresholds, \textit{i.e.}, 0.5: 0.05: 0.95, with all categories. It is the commonly used evaluation metric for object detection.

\subsection{Experimental Setting}

 We utilize ResNet-50, ResNet-101, and ResNeXt-101 with FPN as backbones. The batch normalization layers are fixed to be frozen in the training phase. 
 % We train our model on the MS-COCO 2017 dataset with 135k and 180k iterations for MAL, individually. The base learning rate is set to 0.01 and decreased by a factor of 10 after 90k and 120k for the 135k setting, and 120k and 160k for the 180k setting. 
%  We train our model on the MS-COCO 2017 dataset with 180k iterations. 
We use a mini-batch of 2 images per GPU, thus making a total mini-batch of 16 images on 8 GPUs. 
The initial learning rate is set to 0.01 and decreased by a factor of 10 after 90k and 120k for the 135k setting (ResNet-50), and 120k and 160k for the 180k setting (ResNet-101 and ResNeXt-101).
The synchronized Stochastic Gradient Descent (SGD) is adopted for network optimization. The weight decay of 0.0001 and the momentum of 0.9 are used. A linear warmup strategy is adopted in the first 500 iterations. 
We set the regularization factor $\beta = 0.75$ experimentally. Following \cite{zhang2019freeanchor}, we assign anchors to ground-truth using IoU threshold of 0.5, and to background if their IoUs are in [0, 0.4).

\begin{table*}[t]
    \begin{subtable}[t]{0.27\textwidth}  % 0.29
        \centering
        \fontsize{9}{11}\selectfont
        % \begin{tabular}[t]{c|c|cc}
        \begin{tabular}[t]{c|p{0.52cm} p{0.52cm} p{0.52cm}}
        \hline
        Method  & $\mathrm{AP}$   & $\mathrm{AP_{50}}$ & $\mathrm{AP_{75}}$ \\ \hline \hline
        MAL($k$=40)  & 38.27 & 	56.67 & 	40.81 \\
        MAL($k$=50)  & \textbf{38.39} & \textbf{56.81}  & \textbf{41.14} \\
        MAL($k$=60)  & 38.08 & 56.11 & 40.18 \\\hline
        \end{tabular}
        
        \caption{Detection performance upon different anchor numbers $k$ in each anchor bag.}
        \label{tab:ablation:k}
    \end{subtable}
    \quad
    \begin{subtable}[t]{0.31\textwidth}  % 0.32
        \centering
        \fontsize{9}{11}\selectfont
        % \begin{tabular}[t]{c|c|cc}}
        \begin{tabular}[t]{c|p{0.52cm} p{0.52cm} p{0.52cm}}
        \hline
        Method  & $\mathrm{AP}$   & $\mathrm{AP_{50}}$ & $\mathrm{AP_{75}}$ \\ \hline \hline
        RetinaNet   & 35.46 & 51.61 & 	39.37 \\
        MAL+S(all)  & 38.14 & 	56.81 & 	40.81 \\
        MAL+S(all-top1)  & \textbf{38.39} & 	\textbf{56.81} & 	\textbf{41.14} \\\hline
        \end{tabular}
        \caption{Anchor selection strategy $\phi(\lambda)$. ``S" denotes ``Selection". We compare selecting all instances and all-top1 instance.}
        \label{tab:ablation:selection}
    \end{subtable}
    \quad
    \begin{subtable}[t]{0.37\textwidth}  % 0.38
        \centering
        \fontsize{9}{11}\selectfont
        \begin{tabular}[t]{c|p{0.52cm} p{0.52cm} p{0.52cm}}
        \hline
        Method       & $\mathrm{AP}$   & $\mathrm{AP_{50}}$ & $\mathrm{AP_{75}}$ \\ \hline \hline
        % RetinaNet (baseline) & ResNet-50     & 90k & 35.60 & 	51.89 & 	39.05 & 	19.83 & 	39.94 & 47.42   \\
        MAL+D(constant)  & 35.25 & 	51.72 &	38.92 \\
        MAL+D(step) & 35.88 & 	52.34 & 39.63 \\
        MAL+D(symmetric step)  &  \textbf{36.18} & 	\textbf{52.66} & 	\textbf{39.88}\\ \hline
        \end{tabular}
        \caption{Depression strategy $\psi(\lambda)$. ``D" denotes ``Depression". The constant function, step function, and symmetric step function are compared. }
        \label{tab:ablation:depression}
    \end{subtable}
    \caption{Ablation study on the COCO minval dataset with the backbone ResNet50. We show the AP, AP$_{50}$, and AP$_{75}$ (\%). }
    \label{tab:ablation}
\end{table*} 
\vspace{-1em}
\subsection{Ablation Study}

\begin{figure}[t]
\centering
   \includegraphics[width=\linewidth]{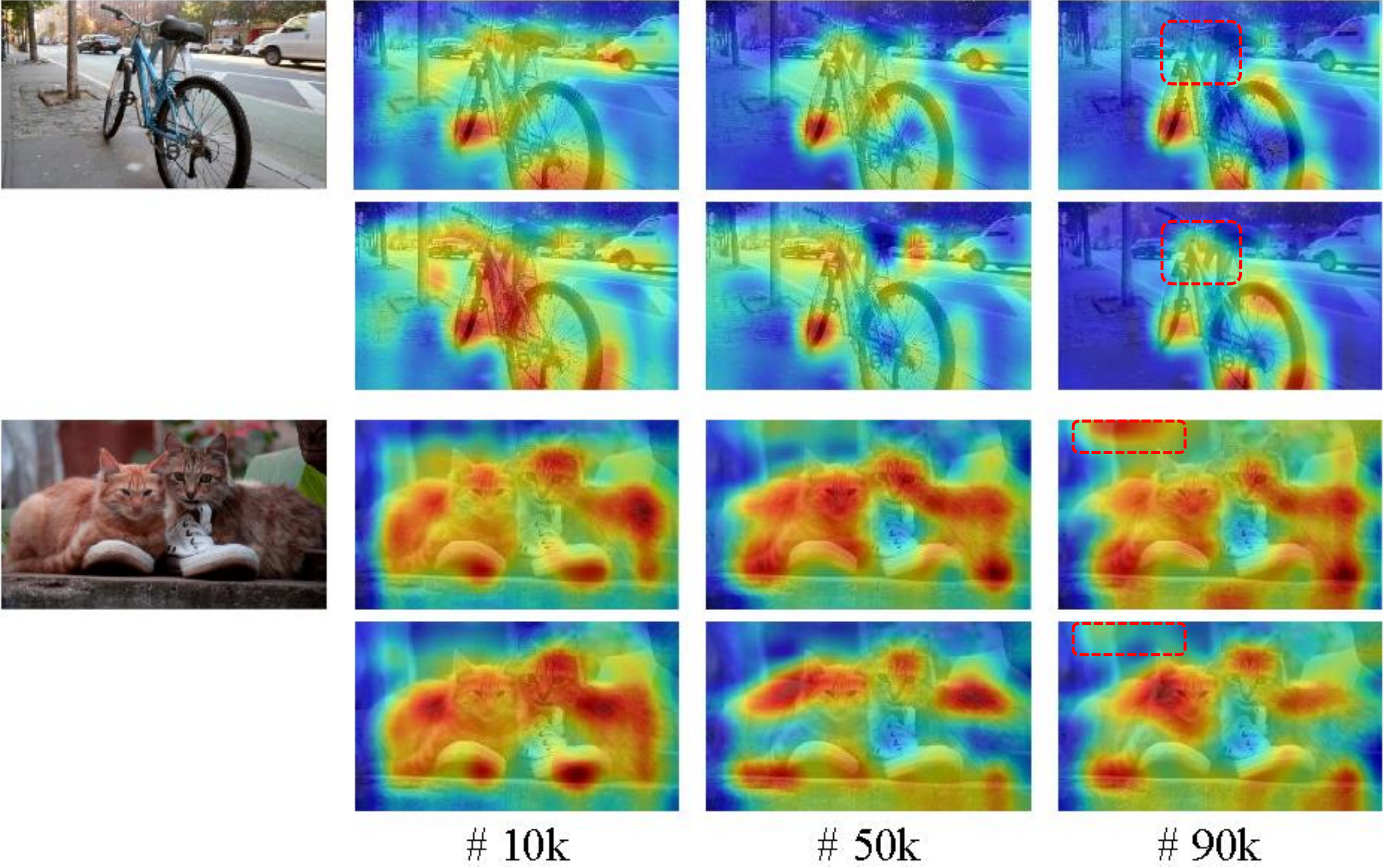}
\caption{The activation map comparison between RetinaNet (the first and third rows) and MAL (the second and fourth rows). The attention maps at the 10k, 50k and 90k iterations are overlaid on input images. As highlighted by red boxes at the 90k $th$ iteration, MAL gets better attention maps which activate more parts in the bicycle image and suppress irrelevant parts in the cat image.  }
\label{fig:effect}
\end{figure}

For ablation study, we used ResNet-50 as the backbone. All detection performances were evaluated on the COCO-minval dataset ($5k$ images). Firstly, we visualize the effectiveness of MAL in Fig. \ref{fig:effect} on feature activation maps. Comparing MAL with RetinaNet, MAL activates more parts on the object and suppresses the more parts in the background. It demonstrates that MAL improved features for better object detection.

\textbf{Anchor Selection}: Without the depression component of MAL, we evaluate the selection component individually first. We compare the results of different $k$ for anchor bag construction, as shown in Table \ref{tab:ablation:k}. The AP is stable when $k=40$, $50$, or $60$. We choose 50 anchors in the following experiments.
% as it achieves the few better performance than others.  
The results of different anchor selection strategies are shown in Table \ref{tab:ablation:selection}.
It improves AP from 35.46\% to 38.14\% when anchor bags are used instead of the scattered anchors in RetinaNet, as MAL+S(all) in Table \ref{tab:ablation:selection}. In the RetinaNet, if an anchor is with good localization but without the highest score, it does not affect the network parameters. While using the anchor bags, this kind of anchor has potential to be selected for detector learning. By the continuation optimization which selects all anchors at the beginning and gradually reduces the selected anchors to the top-1, the performance is further improved to 38.39\%, as MAL+S(all-top1) in Table \ref{tab:ablation:selection}. It verifies that continuation optimization is also efficient in MAL. 

\textbf{Anchor Depression}: We only add the depression component to RetinaNet to find the preferable indicator function $\psi(\lambda)$. We employ three kinds of indicator function. The first one is the constant function, which means keeping the same depression ratio in the whole training phase. We depress the top 50\% pixels in the attention map. The AP decreases a little from 35.46\% to 35.25\%, as MAL+D(constant) shown in Table \ref{tab:ablation:depression}. The reason is that at the beginning of the training phase, the parameters of the network are randomly initialized, and the depression is meaningless for the adversarial learning. 
If a step function is utilized for $\psi(\lambda)$, which increases the depression part from 0.0\% to 50.0\% by step, the performance is increased to 35.88\%, as MAL+D(step) shown in Table \ref{tab:ablation:depression}. It illustrates that the detector should be optimized in a way before depression. The third one is the symmetric step function, which increases the depression part from 0.0\% to 50\% and then decreases it from 50\% to 0.0\%. It achieves the best performance of 36.18\%, as MAL+D(symmetric step) shown in Table \ref{tab:ablation:depression}.

\begin{figure}[t]
\begin{center}
   \includegraphics[width=1.0\linewidth]{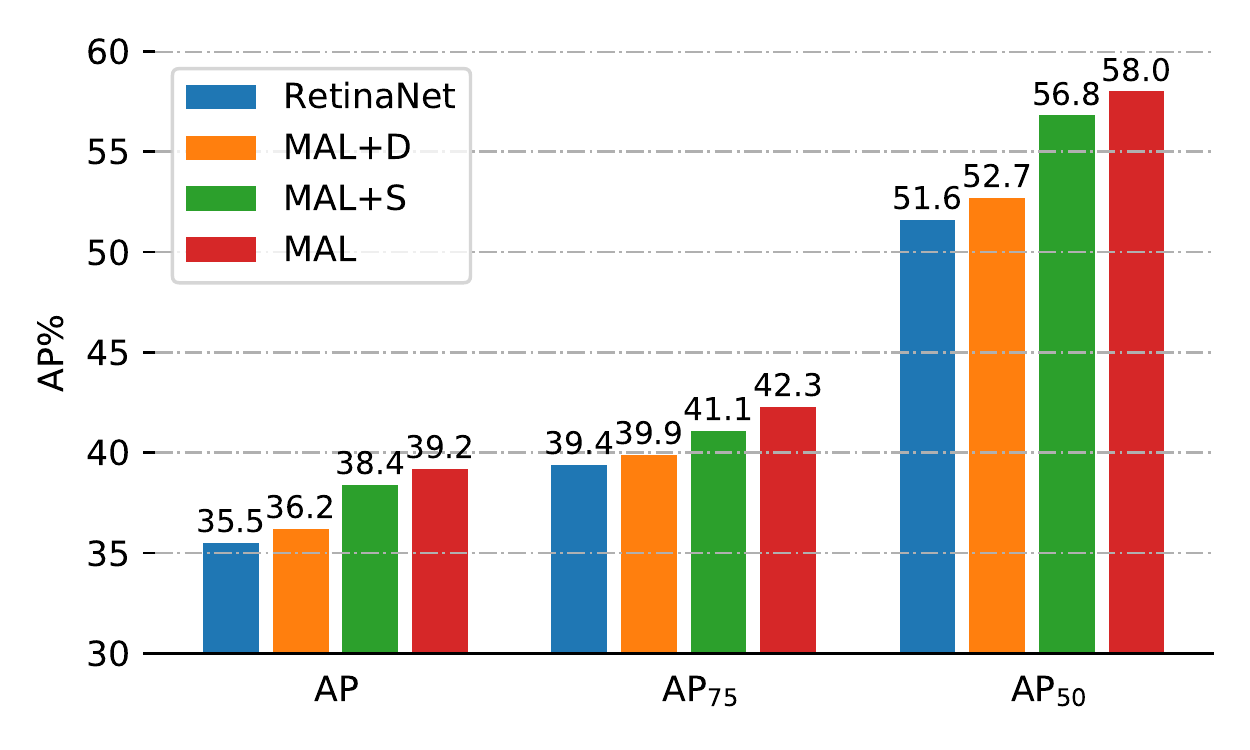}
\end{center}
\vspace{-1em}
\caption{Ablation studies of the anchor selection and depression modules on the COCO-minval dataset. On the metrics AP, $AP_{75}$ and $AP_{50}$, MAL outperforms the baseline detector (RetinaNet) with significant margins. ``S" and ``D" respectively denote ``Selection" and ``Detection".}
\vspace{-1em}
\label{fig:AblationStudies}
\end{figure}

\textbf{Selection-Depression}: The efficient combination of selection and depression is shown in Fig. \ref{fig:AblationStudies}. We compare the AP, AP$_{75}$, and AP$_{50}$. The AP is increased to 36.2\% with the depression component and to 38.4\% with the selection component. When the adversarial manner is taken between selection and the depression, the AP is further improved to 39.2\%, which is 3.7\% (35.5\% vs. 39.2\%) performance gain compared with the original RetinaNet. The AP$_{75}$ and AP$_{50}$ have the same trend of growth as the AP.

\begin{figure}[t]
    \centering
    \begin{subfigure}[b]{0.49\linewidth}
       \includegraphics[width=\linewidth]{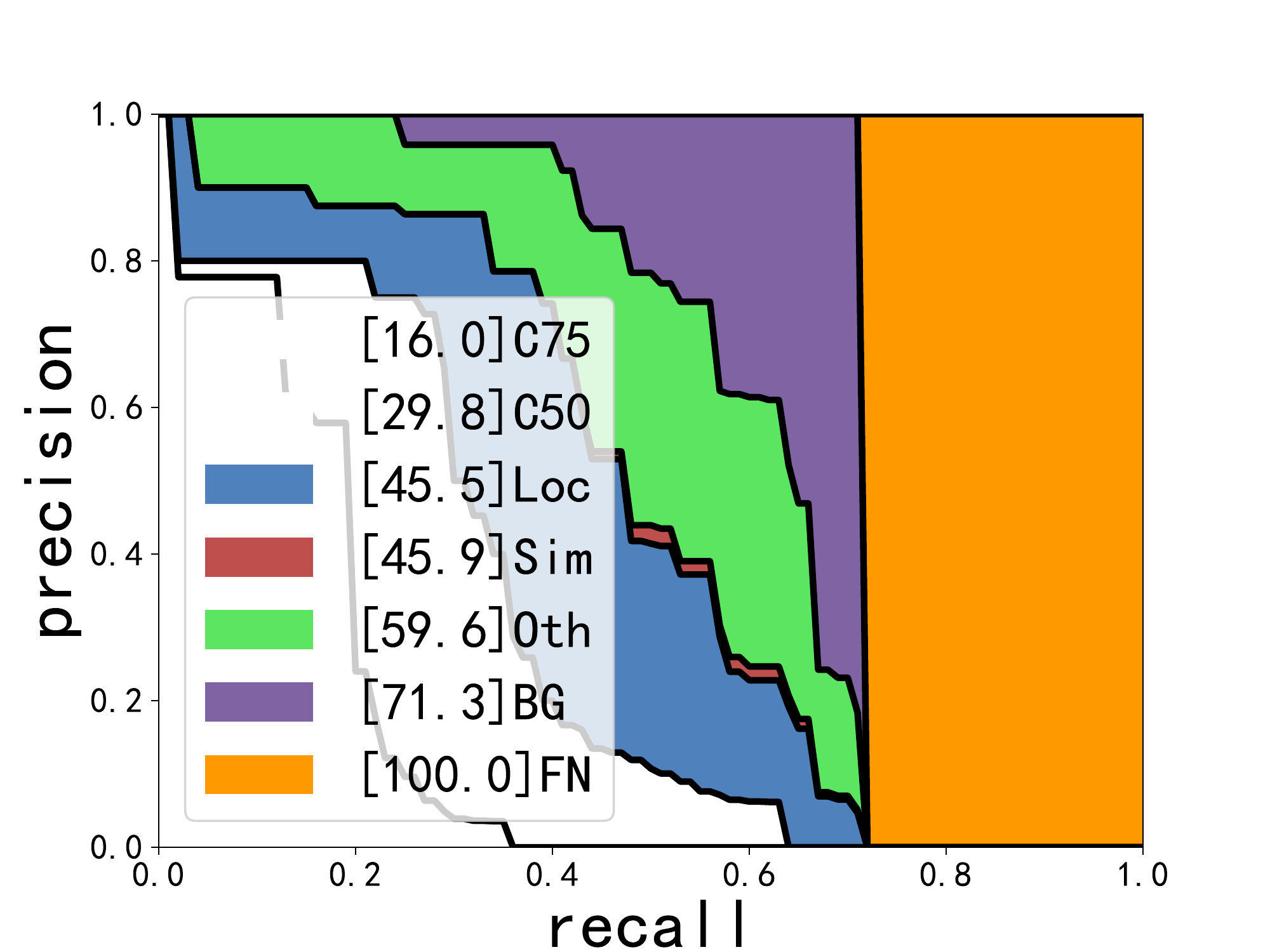}
       \caption{Baseline (label=toothbrush)}
       \label{fig:fig4a}
    \end{subfigure}
    \begin{subfigure}[b]{0.49\linewidth}
       \includegraphics[width=\linewidth]{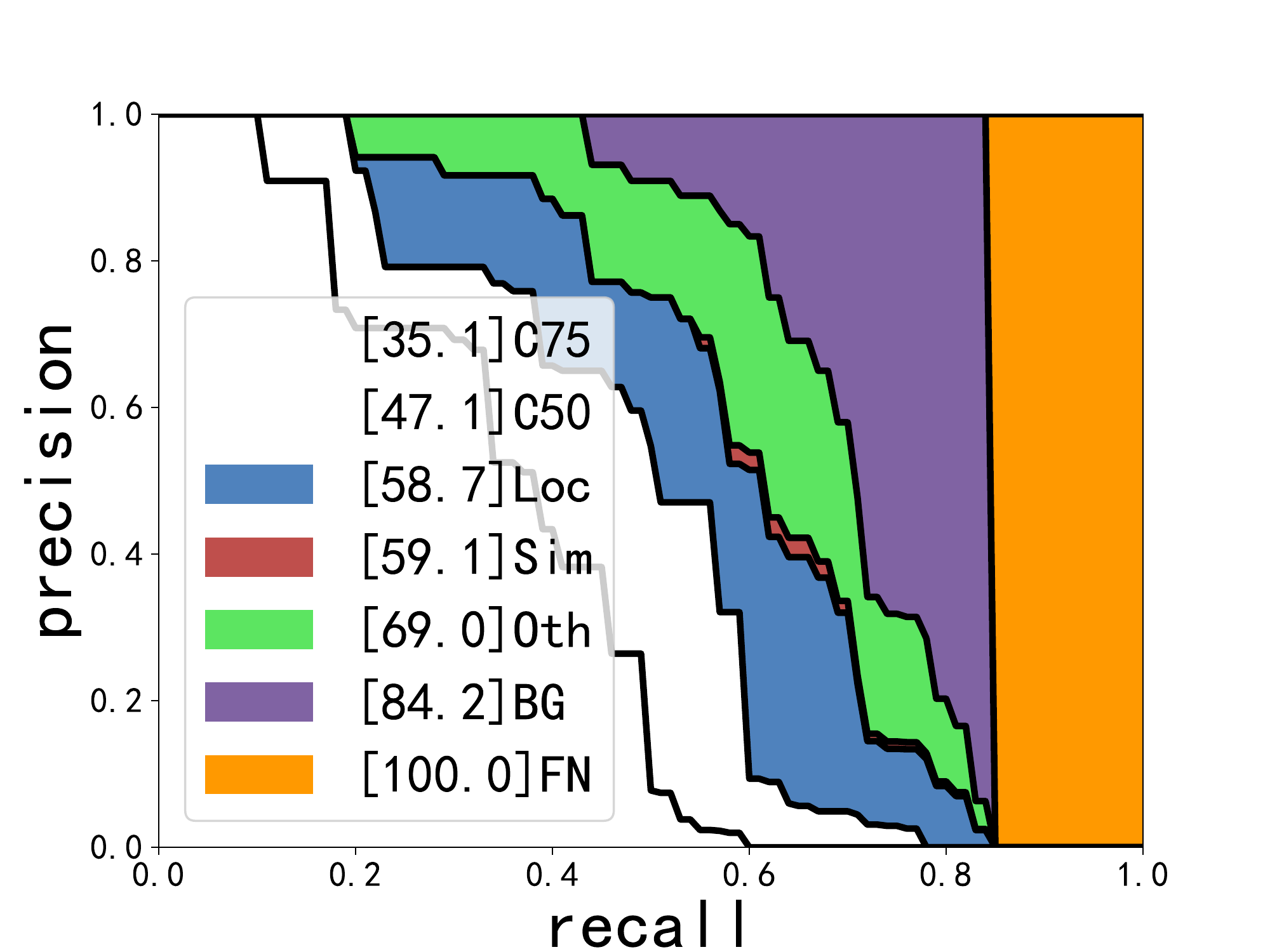}
       \caption{MAL (label=toothbrush)}
       \label{fig:fig4b}
    \end{subfigure}
    \begin{subfigure}[b]{0.49\linewidth}
       \includegraphics[width=\linewidth]{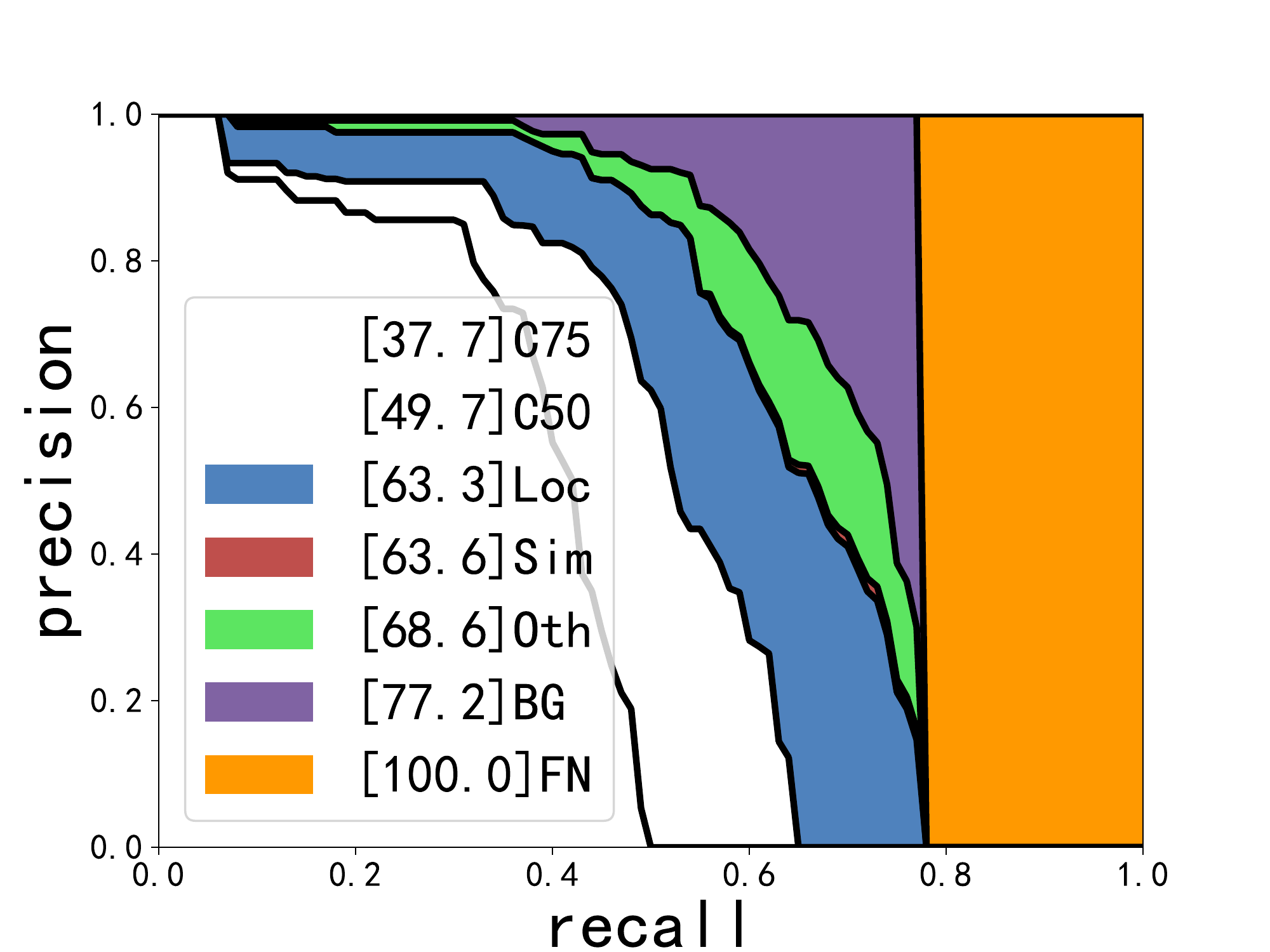}
       \caption{Baseline (label=kite)}
       \label{fig:fig4c}
    \end{subfigure}
    \begin{subfigure}[b]{0.49\linewidth}
       \includegraphics[width=\linewidth]{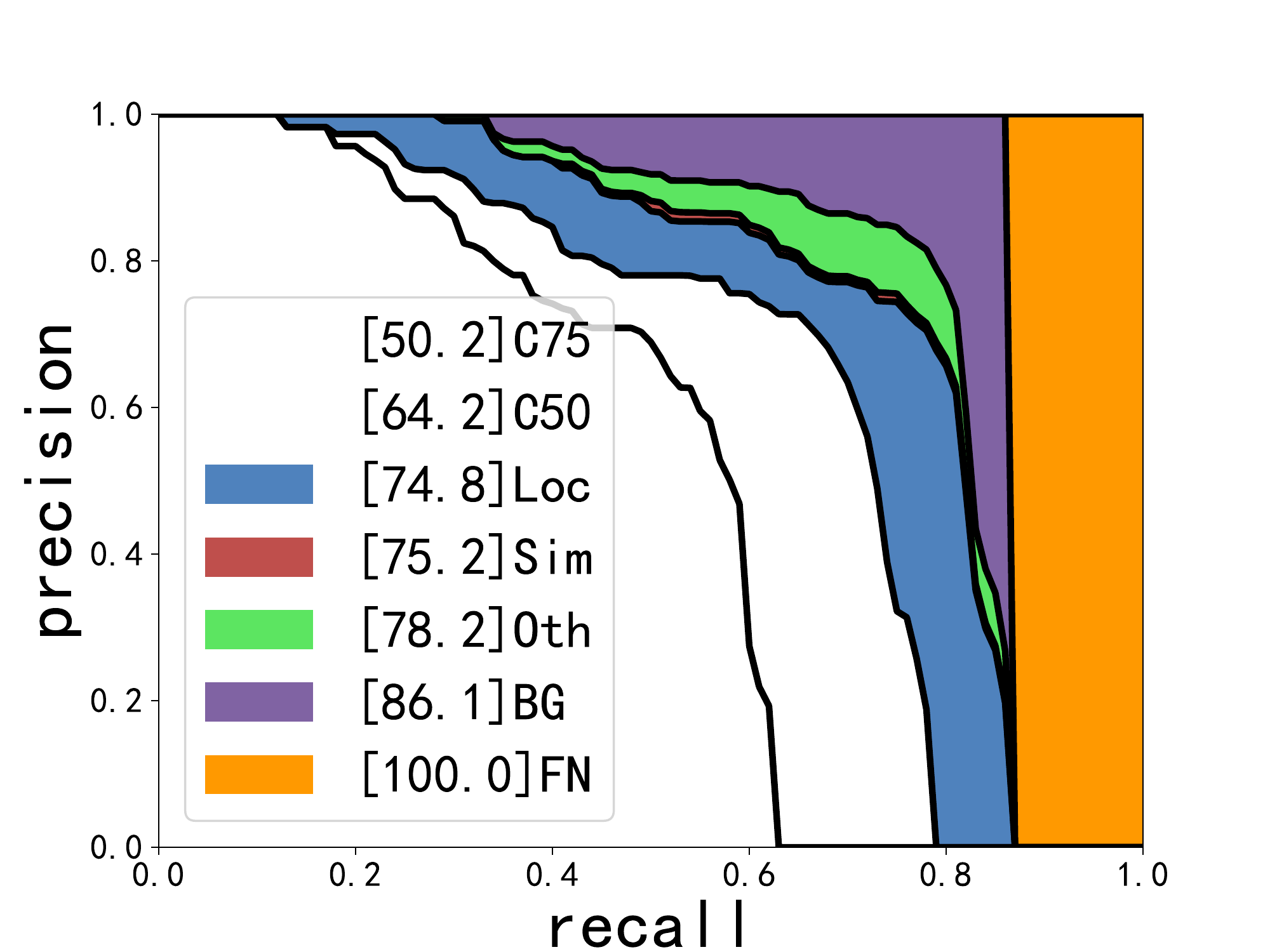}
       \caption{MAL (label=kite)}
       \label{fig:fig4d}
    \end{subfigure}
    \caption{Quantitative evaluation of detection performance. Top row: performance comparison on toothbrush detection. Bottom row: performance summary for the kit category.}
    \label{fig:QuantitativeEvaluation}
\end{figure}

\textbf{Localization Improvement}: 
%The efficient of combination of selection and depression is shown in Fig. 3. We compare the AP, AP$_{75}$, and AP$_{50}$. The AP is increased to 36.2\% with depression component and 38.4\% with selection component. When adversarial manner is token between selection and depression, the AP is further improved to 39.2\%, which is 3.8\% performance gain comparing with the original RetinaNet. The AP$_{75}$ and AP$_{50}$ have the same trend of growth to the AP. 
% add by Tianliang
In Fig. \ref{fig:QuantitativeEvaluation}, we show an error factor analysis~\cite{mmdetection} of the localization results. It can be seen that poor localization (Loc) hinders the improvement of detection performance for objects of irregular shapes, \textit{i.e.}, tilted and slender objects. Compared with the baseline method, MAL significantly reduces the localization error (blue part in Fig. \ref{fig:QuantitativeEvaluation}) of these objects. For instance, the area under curve (AUC) decreases from 15.7\% (45.5\%$-$29.8\%) to 11.6\% (58.7\%$-$47.1\%) for the toothbrush category and from 13.6\% (63.3\%$-$49.7\%) to 10.6\% (74.8\%$-$64.2\%) for the kite category. 

\subsection{Comparison with State-of-the-Art Detectors}

% Table example
% \centering
% \fontsize{7}{9}\selectfont
% Method       & Backbone      & Iteration & $\mathrm{AP}$   & $\mathrm{AP_{50}}$ & $\mathrm{AP_{75}}$ & $\mathrm{AP_S}$  & $\mathrm{AP_{M}}$  & $\mathrm{AP_{L}}$ \\ \hline \hline

\begin{table}[t]
\centering
\fontsize{10}{12}\selectfont
\begin{tabular}{c|c|ccc}
\hline
Method       & Backbone  & $\mathrm{AP}$   & $\mathrm{AP_{50}}$ & $\mathrm{AP_{75}}$  \\ \hline \hline
RetinaNet \cite{FocalLoss17}        & ResNet-50  & 35.5	& 51.6 &	39.4  \\
MAL (ours)        & ResNet-50   & \textbf{39.2}          & \textbf{58.0}          & \textbf{42.3}\\ \hline
% -------------------------------------------------------------------
RetinaNet \cite{FocalLoss17}        & ResNet-101 & 39.1         & 59.1          & 42.3  \\
MAL (ours)        & ResNet-101    & \textbf{43.6} & \textbf{62.8}        & \textbf{47.1}         \\ \hline 
% -------------------------------------------------------------------
RetinaNet \cite{FocalLoss17}        & ResNeXt-101  & 40.8          & 61.1          & 44.1  \\
MAL (ours)        & ResNeXt-101    & \textbf{45.9} & \textbf{65.4} & \textbf{49.7}     \\  \hline
\end{tabular}
\caption{ Performance comparison with the baseline method (single-scale results) on the MS-COCO test-dev dataset. MAL improves the baseline with significant margins.}
\vspace{-1em}
\label{Table_Compared_With_baseline}
\end{table}

\begin{table*}[t]
\centering
\fontsize{10}{12}\selectfont
\begin{tabular}{l|c|cccccc}
\hline
Method       & Backbone       & $\mathrm{AP}$   & $\mathrm{AP_{50}}$ & $\mathrm{AP_{75}}$ & $\mathrm{AP_S}$  & $\mathrm{AP_{M}}$  & $\mathrm{AP_{L}}$ \\ \hline \hline
\textbf{\textit{Two-stage methods}} & & & & & & & \\
Faster R-CNN+++ \cite{ResNet16} & ResNet-101 & 34.9 & 55.7 & 37.4 & 15.6 & 38.7 &  50.9 \\
Faster R-CNN w FPN \cite{FPN17} & ResNet-101 & 36.2 & 59.1 & 39.0 & 18.2 & 39.0 & 48.2 \\
Faster R-CNN w TDM \cite{TDM16} & Inception-ResNet-v2-TDM & 36.8 & 57.7 & 39.2 & 16.2 & 39.8 & 52.1 \\
Deformable R-FCN \cite{dfcn17} & Inception-ResNet-v2 & 37.5 & 58.0 & 40.8 & 19.4 & 40.1 & 52.5 \\
Mask R-CNN \cite{MaskRCNN17}       & ResNeXt-101      & 39.8          & 62.3          & 43.4          & 22.1          & 43.2          & 51.2          \\
IoU-Net \cite{IoU-Net18}          & ResNet-101        & 40.6          & 59.0          & -             & -             & -             & -             \\
Cascade RCNN \cite{cascade18}      & ResNet-101        & 42.8          & 62.1 & 46.3          & 23.7          & 45.5          & 55.2 \\
Grid R-CNN w/ FPN \cite{GridRCNN19} & ResNeXt-101         & 43.2          & 63.0          & 46.6          & 25.1          & 46.5          & 55.2          \\
\hline
\textbf{\textit{One-stage methods}} & & & & & & & \\
YOLOv2 \cite{YOLO16} & DarkNet-19 & 21.6 & 44.0 & 19.2 & 5.0 & 22.4 & 35.5 \\
SSD513 \cite{SSD16} & ResNet-101 & 31.2 & 50.4 & 33.3 & 10.2 & 34.5 & 49.8 \\
YOLOv3 \cite{YOLO9000} & Darknet-53 & 33.0 & 57.9 & 34.4 & 18.3 & 35.4 & 41.9 \\
DSSD513 \cite{SSD16} & ResNet-101 & 33.2 & 53.3 & 35.2 & 13.0 & 35.4 & 51.1 \\
GA-RetinaNet \cite{GuidedAnchoring}      & ResNet-50    & 37.1          & 56.9          & 40.0          & 20.1          & 40.1          & 48.0          \\
MetaAnchor \cite{MetaAnchor2018}       & ResNet-50         & 37.9          & -             & -             & -             & -             & -             \\
RetinaNet \cite{FocalLoss17} &  ResNet101 &  39.1 & 59.1 & 42.3 & 21.8 & 42.7 & 50.2 \\
CornerNet \cite{CornerNet2018}        & Hourglass-104      & 40.6          & 56.4          & 43.2          & 19.1          & 42.8          & 54.3          \\
RetinaNet \cite{FocalLoss17}        & ResNeXt-101      & 40.8          & 61.1          & 44.1          & 24.1          & 44.2          & 51.2          \\
FCOS \cite{tian2019fcos}            & ResNet-101        & 41.5          & 60.7          & 45.0          & 24.4 & 44.8          & 51.6          \\
FoveaBox \cite{kong2019foveabox}         & ResNeXt-101      & 42.1          & 61.9          & 45.2          & 24.9          & 46.8          & 55.6          \\

AB+FSAF \cite{zhu2019feature}          & ResNeXt-101       & 42.9          & 63.8          & 46.3          & 26.6          & 46.2          & 52.7          \\
FreeAnchor \cite{zhang2019freeanchor}       & ResNeXt-101       & 44.8          & 64.3          & 48.4          & 27.0 & 47.9          & 56.0          \\
CenterNet \cite{CenterNet2019}        & Hourglass-104     & 44.9 & 62.4          & 48.1          & 25.6          & 47.4          & 57.4 \\
\hline
\textbf{\textit{ours}} & & & & & & & \\
MAL      & ResNet-101        & 43.6 & 62.8         & 47.1  & 25.0 & 46.9 & 55.8   \\
MAL        & ResNeXt-101       & 45.9 & 65.4 & 49.7 & 27.8 & 49.1 & 57.8 \\ 
MAL (multi-scale) & ResNet-101          & 45.0              &  63.7             & 48.9              & 28.0              & 48.0              & 57.0              \\
MAL (multi-scale) & ResNeXt-101         & \textbf{47.0}             & \textbf{66.1}             & \textbf{51.2}             & \textbf{30.2}             & \textbf{50.1}             & \textbf{58.9}             \\ \hline
\end{tabular}
\vspace{-0.5em}
\caption{ Performance comparison with the state-of-the-art methods on the MS-COCO test-dev dataset (single-scale results unless explicitly stated). MAL achieves new state-of-the-art performance. As a one-stage detector, MAL also outperforms most two-stage detectors. }
\vspace{-0.5em}
\label{Table_Compared_With_SOTA}
\end{table*}

Keeping the best setting in the ablation study, we compare the propsoed MAL with the baseline, $i.e.$, RetinaNet, in Table \ref{Table_Compared_With_baseline}. For ResNet-50, MAL improves the baseline from 35.5\% to 39.2\% with 3.7\% improvement. For ResNet-101 and ResNeXt-101, the improvements are 4.5\% and 4.1\%, respectively. It illustrates that MAL achieves reliable gains with various of backbones. 

In Table \ref{Table_Compared_With_SOTA}, MAL is compared with the state-of-the-art detectors of two-stage methods and one-stage methods on the MS COCO test dataset, which are arranged in the increasing order of AP. For fair comparison, we re-scale the images such that their shorter sides are 800 pixels and the longer sides not more than 1333 pixels.

% including RetinaNet \cite{FocalLoss17}, CornerNet \cite{CornerNet2018}, CenterNet \cite{CenterNet2019}, and FreeAnchor \cite{zhang2019freeanchor}  

%
For one-stage methods, we compare the state-of-the-art including YOLO \cite{YOLO16, YOLO9000}, SSD \cite{SSD16}, FCOS~\cite{tian2019fcos}, FreeAnchor \cite{zhang2019freeanchor} and CenterNet \cite{CenterNet2019}. 
With the ResNet-101 backbone, MAL achieves 43.6\% AP of single-scale, which outperforms the anchor-free approach FCOS~\cite{tian2019fcos} by 2.1\% (43.6\% vs. 41.5\%).
With the ResNeXt-101 backbone, MAL achieves 45.9\% AP of single scale, which achieves 1.1\% (45.9\% vs. 44.8\%) gain compared with the recent FreeAnchor \cite{zhang2019freeanchor}.
It also outperforms state-of-the-art CenterNet \cite{CenterNet2019} by 1.0\% AP (45.9\% vs. 44.9\%). Note that CenterNet uses the Hourglass-104 backbone which has much more network parameters than ResNeXt-101.
These are significant margins for the challenging object detection task. 
The multi-scale testing APs of MAL are further improved to 45.0\% and 47.0\% with ResNet-101 and ResNeXt101, respectively.
% A\P . Compared to recent state-of-the-art one-state detector CornerNet \cite{CornerNet2018}, our MAL achieves a 2.3 AP (42.9 vs. 40.6) improvement. CornerNet uses a heavy backbone (Hourglass-104), it need more computing resource.
% With ResNeXt-101, MAL achieves 45.9\% AP, which outperforms the baseline (RetinaNet) by 5.1\% AP (45.9\% vs. 40.8\%) and outperforms state-of-the-art CenterNet by 1.0\% AP (45.9\% vs. 44.9\%). Note that CenterNet uses the Hourglass-104 backbone which has much more network parameters than ResNeXt-101. In addition, CenterNet uses 480k training iterations which is much more than the 180k we use. Even with much fewer network parameters and iterations, for $\mathrm{AP_{50}}$ and $\mathrm{AP_{75}}$, MAL still outperforms CenterNet by 3.0\% AP and 1.6\% AP, respectively. 

Table \ref{Table_Compared_With_SOTA} also compares MAL with representative two-stage detectors including Faster-RCNN with FPN \cite{FPN17}, Mask R-CNN \cite{MaskRCNN17}, IoU-Net \cite{IoU-Net18}, and Grid R-CNN \cite{GridRCNN19}. MAL outperforms most two-stage detectors. Particularly, it outperforms the recent Grid R-CNN detector by 2.7\% (45.9 vs. 43.2\%) with the same backbone. As a one-stage detector with simpler implementation, MAL shows great potential to surpass two-stage detectors.

%\begin{figure*}[h]
%\begin{center}
%  \includegraphics[width=1.0\linewidth]{latex/pdf/MLA_DetectionResults2.pdf}
%\end{center}
%\caption{Some detections examples. Top: Baseline results by RetinaNet. Bottom: Our results by MAL.}
%\end{figure*}

\section{Conclusion}
We have proposed an elegant and effective training approach, referred to as Multiple Anchor Learning (MAL), for visual object detection. By selecting anchors to jointly optimize bounding box classification and localization, MAL upgrades the standard hand-crafted anchor assignment mechanism to a learnable object-anchor matching mechanism. We proposed a simple selection-depression strategy to alleviate the sub-optimization issue of MAL. MAL improved object detection with significant margins compared with the baseline detector RetinaNet, achieves the best result on MS-COCO among single-stage methods, and outperforms many recent two-stage approaches. Such improvements root in not only the optimal selection of anchors but also implicit feature assembling based on a bag of anchors. Our work presents a promising direction to relax anchor design in learning a visual object detection. 
% \newpage 

{\small
\bibliographystyle{ieee_fullname}
\bibliography{egbib}

\begin{thebibliography}{10}\itemsep=-1pt

\bibitem{cascade18}
Zhaowei Cai and Nuno Vasconcelos.
\newblock Cascade r-cnn: Delving into high quality object detection.
\newblock In {\em IEEE CVPR}, pages 6154--6162, 2018.

\bibitem{mmdetection}
Kai Chen, Jiaqi Wang, Jiangmiao Pang, Yuhang Cao, Yu Xiong, Xiaoxiao Li,
  Shuyang Sun, Wansen Feng, Ziwei Liu, Jiarui Xu, Zheng Zhang, Dazhi Cheng,
  Chenchen Zhu, Tianheng Cheng, Qijie Zhao, Buyu Li, Xin Lu, Rui Zhu, Yue Wu,
  Jifeng Dai, Jingdong Wang, Jianping Shi, Wanli Ouyang, Chen~Change Loy, and
  Dahua Lin.
\newblock {MMDetection}: Open mmlab detection toolbox and benchmark.
\newblock {\em arXiv:1906.07155}, 2019.

\bibitem{GaussianYOLO2019}
Jiwoong Choi, Dayoung Chun, Hyun Kim, and Hyuk-Jae Lee.
\newblock Gaussian yolov3: An accurate and fast object detector using
  localization uncertainty for autonomous driving.
\newblock In {\em IEEE ICCV}, pages 502--511, 2019.

\bibitem{dfcn17}
Jifeng Dai, Haozhi Qi, Yuwen Xiong, Yi Li, Guodong Zhang, Han Hu, and Yichen
  Wei.
\newblock Deformable convolutional networks.
\newblock In {\em IEEE ICCV}, page 764–773, 2017.

\bibitem{CenterNet2019}
Kaiwen Duan, Song Bai, Lingxi Xie, Honggang Qi, Qingming Huang, and Qi Tian.
\newblock Centernet: Object detection with keypoint triplets.
\newblock In {\em IEEE CVPR}, 2019.

\bibitem{Fu2016dssd}
Cheng-Yang Fu, Wei Liu, Ananth Ranga, Ambrish Tyagi, and Alexander~C. Berg.
\newblock {DSSD}: Deconvolutional single shot detector.
\newblock In {\em arXiv preprint arXiv:1701.06659}.

\bibitem{NAS-FPN2019}
Golnaz Ghiasi, Tsung{-}Yi Lin, and Quoc~V. Le.
\newblock {NAS-FPN:} learning scalable feature pyramid architecture for object
  detection.
\newblock In {\em IEEE CVPR}, pages 7036--7045, 2019.

\bibitem{FastRCNN15}
Ross~B. Girshick.
\newblock Fast {R-CNN}.
\newblock In {\em IEEE ICCV}, pages 1440--1448, 2015.

\bibitem{RCNN14}
Ross~B. Girshick, Jeff Donahue, Trevor Darrell, and Jitendra Malik.
\newblock Rich feature hierarchies for accurate object detection and semantic
  segmentation.
\newblock In {\em IEEE CVPR}, pages 580--587, 2014.

\bibitem{MaskRCNN17}
Kaiming He, Georgia Gkioxari, Piotr Doll{\'{a}}r, and Ross~B. Girshick.
\newblock Mask {R-CNN}.
\newblock In {\em IEEE ICCV}, pages 2980--2988, 2017.

\bibitem{ResNet16}
Kaiming He, Xiangyu Zhang, Shaoqing Ren, and Jian Sun.
\newblock Deep residual learning for image recognition.
\newblock In {\em IEEE CVPR}, pages 770--778, 2016.

\bibitem{hzeIAN2019}
Zeyi Huang, Wei Ke, and Dong Huang.
\newblock Improving object detection with inverted attention.
\newblock {\em arXiv:1903.12255}, 2019.

\bibitem{IoU-Net18}
Borui Jiang, Ruixuan Luo, Jiayuan Mao, Tete Xiao, and Yuning Jiang.
\newblock Acquisition of localization confidence for accurate object detection.
\newblock In {\em ECCV}, pages 784--799, 2018.

\bibitem{kong2019foveabox}
Tao Kong, Fuchun Sun, Huaping Liu, Yuning Jiang, and Jianbo Shi.
\newblock Foveabox: Beyond anchor-based object detector.
\newblock {\em arXiv:1904.03797}, 2019.

\bibitem{CornerNet2018}
Hei Law and Jia Deng.
\newblock Cornernet: Detecting objects as paired keypoints.
\newblock In {\em ECCV}, pages 765--781, 2018.

\bibitem{Triplet2019}
Yanghao Li, Yuntao Chen, Naiyan Wang, and Zhaoxiang Zhang.
\newblock Scale-aware trident networks for object detection.
\newblock In {\em IEEE ICCV}, pages 502--511, 2019.

\bibitem{FPN17}
Tsung{-}Yi Lin, Piotr Doll{\'{a}}r, Ross~B. Girshick, Kaiming He, Bharath
  Hariharan, and Serge~J. Belongie.
\newblock Feature pyramid networks for object detection.
\newblock In {\em IEEE CVPR}, pages 936--944, 2017.

\bibitem{FocalLoss17}
Tsung{-}Yi Lin, Priya Goyal, Ross~B. Girshick, Kaiming He, and Piotr
  Doll{\'{a}}r.
\newblock Focal loss for dense object detection.
\newblock In {\em IEEE ICCV}, pages 2999--3007, 2017.

\bibitem{DBLP:conf/eccv/LinMBHPRDZ14}
Tsung{-}Yi Lin, Michael Maire, Serge~J. Belongie, James Hays, Pietro Perona,
  Deva Ramanan, Piotr Doll{\'{a}}r, and C.~Lawrence Zitnick.
\newblock Microsoft {COCO:} common objects in context.
\newblock In {\em ECCV}, 2014.

\bibitem{Survey2019}
Li Liu, Wanli Ouyang, XiaogangWang, Paul Fieguth, Jie Chen, Xinwang Liu, and
  Matti Pietikainen.
\newblock Deep learning for generic object detection: A survey.
\newblock {\em Int. J. Comp. Vis.}, 2019.

\bibitem{SSD16}
Wei Liu, Dragomir Anguelov, Dumitru Erhan, Christian Szegedy, Scott~E. Reed,
  Cheng{-}Yang Fu, and Alexander~C. Berg.
\newblock {SSD:} single shot multibox detector.
\newblock In {\em ECCV}, pages 21--37, 2016.

\bibitem{GridRCNN19}
Xin Lu, Buyu Li, Yuxin Yue, Quanquan Li, and Junjie Yan.
\newblock Grid {R-CNN}.
\newblock In {\em IEEE CVPR}, 2019.

\bibitem{MIL97}
Oded Maron and Tom{\'{a}}s Lozano{-}P{\'{e}}rez.
\newblock A framework for multiple-instance learning.
\newblock In {\em NeurIPS}, pages 570--576, 1997.

\bibitem{ScaleSensitive2019}
Junran Peng, Ming Sun, Zhaoxiang Zhang, Tieniu Tan, and Junjie Yan.
\newblock Pod: Practical object detection with scale-sensitive network.
\newblock In {\em IEEE ICCV}, pages 6054--6063, 2019.

\bibitem{YOLO16}
Joseph Redmon, Santosh~Kumar Divvala, Ross~B. Girshick, and Ali Farhadi.
\newblock You only look once: Unified, real-time object detection.
\newblock In {\em IEEE CVPR}, pages 779--788, 2016.

\bibitem{YOLO9000}
Joseph Redmon and Ali Farhadi.
\newblock {YOLO9000:} better, faster, stronger.
\newblock In {\em IEEE CVPR}, pages 6517--6525, 2017.

\bibitem{FasterRCNN15}
Shaoqing Ren, Kaiming He, Ross~B. Girshick, and Jian Sun.
\newblock Faster {R-CNN:} towards real-time object detection with region
  proposal networks.
\newblock In {\em NeurIPS}, pages 91--99, 2015.

\bibitem{TDM16}
Abhinav Shrivastava, Rahul Sukthankar, Jitendra Malik, and Abhinav Gupta.
\newblock Beyond skip connections: Top-down modulation for object detection.
\newblock {\em arXiv:1612.06851}, 2016.

\bibitem{tian2019fcos}
Zhi Tian, Chunhua Shen, Hao Chen, and Tong He.
\newblock Fcos: Fully convolutional one-stage object detection.
\newblock {\em arXiv:1904.01355}, 2019.

\bibitem{cvpr_WanLKJJY19}
Fang Wan, Chang Liu, Wei Ke, Xiangyang Ji, Jianbin Jiao, and Qixiang Ye.
\newblock {C-MIL:} continuation multiple instance learning for weakly
  supervised object detection.
\newblock In {\em IEEE CVPR}, pages 2199--2208, 2019.

\bibitem{GuidedAnchoring}
Jiaqi Wang, Kai Chen, Shuo Yang, Chen~Change Loy, and Dahua Lin.
\newblock Region proposal by guided anchoring.
\newblock In {\em IEEE CVPR}, pages 2965--2974, 2019.

\bibitem{MetaAnchor2018}
Tong Yang, Xiangyu Zhang, Zeming Li, Wenqiang Zhang, and Jian Sun.
\newblock Metaanchor: Learning to detect objects with customized anchors.
\newblock In {\em NeurIPS}, pages 320--330, 2018.

\bibitem{RepPoint2019}
Ze Yang, Shaohui Liu, Han Hu, Liwei Wang, and Stephen Lin.
\newblock Reppoints: Point set representation for object detection.
\newblock In {\em IEEE ICCV}, pages 502--511, 2019.

\bibitem{zhang2019freeanchor}
Xiaosong Zhang, Fang Wan, Chang Liu, Rongrong Ji, and Qixiang Ye.
\newblock {FreeAnchor}: Learning to match anchors for visual object detection.
\newblock In {\em NeurIPS}, 2019.

\bibitem{ExremePoint2019}
Xingyi Zhou, Jiacheng Zhuo, and Philipp Kr{\"{a}}henb{\"{u}}hl.
\newblock Bottom-up object detection by grouping extreme and center points.
\newblock In {\em IEEE CVPR}, pages 850--859, 2019.

\bibitem{zhu2019feature}
Chenchen Zhu, Yihui He, and Marios Savvides.
\newblock Feature selective anchor-free module for single-shot object
  detection.
\newblock In {\em IEEE CVPR}, pages 840--849, 2019.

\end{thebibliography}
}

\end{document}